\documentclass{article}
\usepackage{spconf,amsmath}

\input{math_commands.sty}
\usepackage{booktabs}
\usepackage[ruled]{algorithm}
\usepackage{algpseudocode}
\title{POLICE: Provably Optimal Linear Constraint Enforcement for\\Deep Neural Networks}
\twoauthors
 {Randall Balestriero}
	{Meta AI, FAIR\\
	\texttt{rbalestriero@meta.com}\\
	NYC, USA}
 {Yann LeCun}
	{Meta AI, FAIR, NYU\\
	\texttt{yann@meta.com}\\
	NYC, USA}
\begin{document}
%\ninept
%
\maketitle

\begin{abstract}
Deep Neural Networks (DNNs) outshine alternative function approximators in many settings thanks to their modularity in composing any desired differentiable operator. The formed parametrized functional is then tuned to solve a task at hand from simple gradient descent. This modularity comes at the cost of making strict enforcement of constraints on DNNs, e.g. from a priori knowledge of the task, or from desired physical properties, an open challenge. In this paper we propose the first provable affine constraint enforcement method for DNNs that only requires minimal changes into a given DNN's forward-pass, that is computationally friendly, and that leaves the optimization of the DNN's parameter to be unconstrained, i.e. standard gradient-based method can be employed. 
Our method does not require any sampling and provably ensures that the DNN fulfills the affine constraint on a given input space's region at any point during training, and testing. We coin this method POLICE, standing for {\bf P}rovably {\bf O}ptimal {\bf LI}near {\bf C}onstraint {\bf E}nforcement.
Github:{\footnotesize 
\noindent\url{https://github.com/RandallBalestriero/POLICE}}
\end{abstract}

\vspace{-0.2cm}
\section{Introduction}
\vspace{-0.2cm}

Deep Neural Networks (DNNs) are compositions of interleaved linear and nonlinear operators forming a differentiable functional $f_{\vtheta}$ governed by some parameters $\vtheta$ \cite{lecun2015deep}. DNNs are universal approximators, but so are many other alternatives e.g. Fourier series \cite{bracewell1986fourier}, kernel machines \cite{park1991universal}, and decision trees \cite{mohri2018foundations}. The strength of DNNs rather lies in the ability of the function $f_{\vtheta}$ to quickly and relatively easily fit most encountered target functional $f^*$ --often known through a given and finite training set-- only by performing gradient updates on its parameters $\vtheta$ \cite{le1991eigenvalues}. This success, combined with the development of software and hardware making DNNs' training and inference efficient, have made DNNs the operator of choice for most tasks encountered in machine learning and pattern recognition \cite{goodfellow2016deep}.

\begin{figure}[t!]
    \centering
    \includegraphics[width=\linewidth]{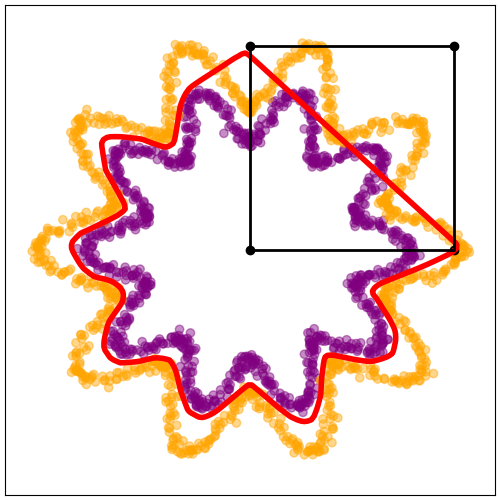}
    \vspace{-0.9cm}
    \caption{\small Classification task of \color{orange}{\bf orange} \color{black} versus \color{purple} {\bf purple} \color{black}. The learned decision boundary (\color{red} {\bf red}\color{black}) comes from a SGD trained {\bf POLICE}d DNN of depth $3$ with leaky-ReLU and dense layers of width $256$; POLICE provably enforce the DNN to be affine within the region delimited by the {\bf black} lines through a simple projection of each layer's bias parameters (\cref{algo:police}) --standard unconstrained optimization is still be used to train the DNN. See \cref{fig:cases} for regression tasks.}
    \label{fig:classification}
    \vspace{-0.4cm}
\end{figure}

Although ubiquitous for general purpose function fitting, DNNs have had a much harder time to shine in settings where specific constraints must be enforced exactly on the learned mapping. A motivating example follows from the following optimization problem
\begin{align}
    \min_{\vtheta} \|f_{\vtheta} -f^* \|+ \lambda \mathcal{R}(\vtheta) \nonumber\\
    \text{ s.t. } \mJ f_{\vtheta}(\vx) = \mA, \forall \vx \in R,\label{eq:constraint}
\end{align}
where $\mJ$ is the Jacobian operator and $\mA$ is a target matrix that the model $f_{\vtheta}$'s Jacobian should match for any input $\vx$ in a given region $R$ of the input space. We employed explicitly a regularizer $\mathcal{R}$ with associated hyper-parameter $\lambda$ as one commonly employs weight-decay i.e. $\ell_2$ regularization on $\vtheta$ \cite{krogh1991simple}. The type of constraint of \cref{eq:constraint} is crucial in a variety of applications e.g. based on informed physical properties that $f_{\vtheta}$ should obey \cite{bers1964partial,farlow1993partial,evans2010partial} or based on (adversarial) noise robustness constraints \cite{goodfellow2014explaining,yuan2019adversarial}. Another motivating example emerges when one has a priori knowledge that the target mapping $f^*$ is itself linear with parameters $\mA^*,\vb^*$ within a region $R$ leading to the following optimization problem
\begin{align}
    \min_{\vtheta} \|f_{\vtheta} -f^* \|+ \lambda \mathcal{R}(\vtheta) \nonumber\\
    \text{ s.t. } f_{\vtheta}(\vx) = \mA^*\vx+\vb^*, \forall \vx \in R.\label{eq:constraint2}
\end{align}
It is easy to see that one might arrive at numerous variants of \cref{eq:constraint,eq:constraint2}'s constraints. Crucial to our study is the observation that in all cases {\em the necessary constraint one must provably and strictly enforce into $f_{\vtheta}$ is to be affine within $R$}. As a result, we summarize the generalized affine constraint that will interest us as follows
\begin{align}
    \min_{\vtheta} \|f_{\vtheta} -f^* \|+ \lambda \mathcal{R}(\vtheta) \nonumber\\
    \text{ s.t. } f_{\vtheta}(\vx) \text{ is affine on $R$},\label{eq:constraint3}
\end{align}
and from which any more specific constraint can be efficiently enforced. For example, given \cref{eq:constraint3}, imposing \cref{eq:constraint} is straightforward by using a barrier method \cite{forsgren2002interior} with $\|\mJ f_{\vtheta}(\vv) - \mA \|$ which can be done only by sampling a single sample $\vv \in R$ and computing the DNN's Jacobian matrix at that point. The same goes for imposing \cref{eq:constraint2} given that \cref{eq:constraint3} is fulfilled.

The main question that naturally arises is how to leverage DNNs while imposing \cref{eq:constraint3}. Current constraint enforcement with DNNs have so far only considered questions of verification \cite{sun2018concolic,wong2018provable,liu2021algorithms} or constrained parameter space e.g. with integer parameters $\vtheta$ \cite{wu2018training} leaving us with little help to solve \cref{eq:constraint3}.
A direct regularization-based approach suffers from the curse of dimensionality \cite{bellman2015applied,koppen2000curse} since sampling $R$ to ensure that $f_{\vtheta}$ is affine requires exponentially many points as the dimension of $R$ increases. A more direct approach e.g. defining an alternative mapping $g_{\vtheta,\mA,\vb}$ such as
\begin{align*}
    g_{\vtheta,\mA,\vb}(\vx)=1_{\{\vx \in R\}} (\mA\vx+\vb)+ 1_{\{\vx \not \in R\}}f_{\vtheta}(\vx),
\end{align*}
would destroy key properties such as continuity at the boundary $\partial R$. In short, {\em we are in need of a principled and informed solution to strictly enforce \cref{eq:constraint3} in arbitrarily wide and/or deep DNNs}.

In this paper, we propose a theoretically motivated and provably optimal solution to enforce the affine constraint of \cref{eq:constraint3} into DNNs. In particular, by focusing on convex polytopal regions $R$, we are able to exactly enforce the model to be affine on $R$; a visual depiction of the method at work is given in \cref{fig:classification}, and the pseudo-code is given in \cref{algo:police}. Our method has linear asymptotic complexity with respect to the number of vertices $P$ defining (or approximating) the region $R$ i.e. it has time-complexity $\mathcal{O}(KP)$ where $K$ is the complexity of the model's forward pass. Hence the proposed strategy will often be unimpaired by the curse of dimensionality since for example a simplicial region $R \subset \mathbb{R}^{D}$ has only $D+1$ vertices \cite{edelsbrunner1987algorithms} i.e. our method in this case has complexity growing linearly with the input space dimension. Our method is exact for any DNN architecture in which the nonlinearities are such as (leaky-)ReLU, absolute value, max-pooling and the likes, extension to smooth nonlinearities is discussed in \cref{sec:conclusion}. We coin our method {\bf POLICE} standing for {\bf P}rovably {\bf O}ptimal {\bf LI}near {\bf C}onstraint {\bf E}nforcement; and we denoted a constrained DNN as being {\bf POLICE}d. Two crucial benefits of POLICE are that the constraint enforcement is exact without requiring any sampling, and that the standard gradient-based training can be used on the POLICEd DNN parameters without any changes.

\vspace{-0.3cm}
\section{Strict Enforcement of Affine Constraints for Deep Neural Networks}
\vspace{-0.2cm}

In this section, we develop the proposed POLICE method that relies on two ingredients: (i) a DNN $f_{\vtheta}$ using continuous piecewise-affine (CPA) nonlinearities which we formally introduce in \cref{sec:MASO}; and (ii) a convex polytopal region $R$. The derivations of the proposed method will be carried in \cref{sec:POLICE} and will empirically validated in \cref{sec:validation}.

\vspace{-0.2cm}
\subsection{Deep Neural Networks Are Continuous Piecewise Linear Operators}
\label{sec:MASO}
\vspace{-0.1cm}

We denote the DNN input-output mapping as $f_{\vtheta}:\mathbb{R}^D \mapsto \mathbb{R}^K$ with $\vtheta$ the governing parameters of the mapping. In all generality, a DNN is formed by a composition of many {\em layers} as in
\begin{align}
    f_{\vtheta}=\left(f_{\vtheta^{(L)}}^{(L)} \circ \dots \circ f_{\vtheta^{(1)}}^{(1)}\right),\label{eq:DNN}
\end{align}
where each layer mapping $f^{(\ell)}: \mathbb{R}^{D^{(\ell)}}\mapsto \mathbb{R}^{D^{(\ell+1)}}$ produces a {\em feature map}; with $D^{(1)}\triangleq D$ and $D^{(L)}\triangleq K$. For most architectures, i.e. parametrizations of the layers, the input-output mapping producing each feature map takes the form 
\begin{align}
    f_{\vtheta^{(\ell)}}^{(\ell)}(\vv)=\sigma^{(\ell)}(\vh^{(\ell)}(\vv))\text{ with } \vh^{(\ell)}(\vv)=\mW^{(\ell)}\vv+\vb^{(\ell)}\label{eq:layer}
\end{align}
where $\sigma$ is a pointwise activation function, $\mW^{(\ell)}$ is a weight matrix of dimensions $D^{(\ell+1)} \times D^{(\ell)}$, and $\vb^{(\ell)}$ is a bias vector of length $D^{(\ell+1)}$, $\vh^{(\ell)}$ is denoted as the {\em pre-activation map}, and the layer parameters are gathered into $\vtheta^{(\ell)} \triangleq \{ \mW^{(\ell)},\vb^{(\ell)}\}$. 
The matrix $\mW^{(\ell)}$ will often have a specific structure (e.g., circulant) at different layers. Without loss of generality we consider vectors as inputs since when dealing with images for example, one can always flatten them into vectors and reparametrize the layers accordingly leaving the input-output mapping of the DNN unchanged. The details on the layer mapping will not impact our results. One recent line of research that we will heavily rely on consists in formulating DNNs as Continuous Piecewise Affine (CPA) mappings \cite{montufar2014number,balestriero2018spline}, that be expressed as
\begin{align}
    f_{\vtheta}(\vx) = \sum_{\omega \in \Omega}(\mA_{\omega}\vx+\vb_{\omega})1_{\{\vx \in \omega\}},\label{eq:CPA}
\end{align}
where $\Omega$ is the input space partition induced by the DNN architecture\footnote{exact relations between architecture and partition are given in \cite{balestriero2019geometry} and are beyond the scope of this study}, $\omega$ is a partition-region, and $\mA_{\omega}, \vb_{\omega}$ are the corresponding per-region slope and offset parameters. A Key result that we will build upon is that the CPA formulation of \cref{eq:CPA} represents exactly the DNN functional of \cref{eq:DNN,eq:layer}, when the nonlinearities $\sigma$ are themselves CPA e.g. (leaky-)ReLU, absolute value, max-pooling \cite{balestriero2018spline}.

\vspace{-0.2cm}
\subsection{POLICE Algorithm and Optimality}
\label{sec:POLICE}
\vspace{-0.1cm}

Equipped with the CPA notations expressing DNNs as per-region affine mappings (recall \cref{eq:CPA}) we now derive the proposed POLICE algorithm that will strictly enforce the POLICEd DNN to fulfill the affine constraint from \cref{eq:constraint3}) on a region $R$ which is convex. We will then empirically validate the method in the following \cref{sec:validation}.

As mentioned above, our goal is to enforce the DNN to be a simple affine mapping on a convex region $R$. In particular, we will consider the $V$-representation of the region $R$ i.e. we consider $P$ vertices $\vv_1,\dots,\vv_{P}$ so that the considered region $R$ is, or can be closely approximated by, the convex hull of those vertices as in
\begin{align}
    R = \left\{\mV^T\valpha: \valpha_p \geq 0, \forall p \wedge  \valpha^{T}\mathbf{1}_{P}=1\right\},\label{eq:convex}
\end{align}
where we gathered the $P$ vertices into the $P \times D$ matrix
\begin{align}
 \mV\triangleq [\vv_{1},\dots,\vv_{P}]^T.\label{eq:V}
\end{align}
We highlight that the actual ordering of the rows of $\mV$ in \cref{eq:V}, i.e. which vertex is put in which row, will not impact our result.
We will also denote by $\vv^{(\ell)}_p$ the $p^{\rm th}$ vertex mapped through the first $\ell-1$ layer, starting with $\vv_p$ for $\ell=1$ and the corresponding matrix $\mV^{(\ell)}$. By leveraging the CPA property of the DNN $f_{\vtheta}$, one should notice that a necessary and sufficient condition for the model to stay affine within $R$ is to have the same pre-activation sign patterns (recall \cref{eq:layer}) which we formalize below. 

\vspace{-0.1cm}
\begin{theorem}
A necessary and sufficient condition for a CPA DNN as per \cref{eq:CPA} to stay affine with a region $R$ given by \cref{eq:convex} is to have the same pre-activation sign patterns between the vertices $\vv_p,\forall p\in[P]$ and that for each layer.
\label{thm:police}
\end{theorem}
\vspace{-0.1cm}

\begin{proof}
The proof is direct. If the pre-activation signs are the same for all vertices, e.g. ${\rm sign}(\vh^{(1)}(\vv_p))={\rm sign}(\vh^{(1)}(\vv_q))$, $\forall p,q \in [P]^2$ for the first layer, then the vertices all lie within the same partition region i.e. $\exists \omega^* \in \Omega $ such that $\vv_p \in \omega, \forall p$. Using \cref{eq:convex} we obtain that $R \subset \omega^*$.
Now, because the partition regions $\omega$ that form the DNN partition $\Omega$ (recall \cref{eq:CPA}) are themselves convex polytopes, it implies that all the points within $\omega^*$, and thus $R$ since we showed that $R \subset \omega^*$, will also have the same pre-activation patterns, i.e. the entire DNN input-output mapping will stay affine within $R$ concluding the proof.
\end{proof}

\cref{thm:police} provides us with a necessary and sufficient condition but does not yet answer the principal question of ``how to ensure that the vertices have the same pre-activation sign patterns''. There are actually many ways to enforce this, in our study we focus on one method that we deem the simplest and leave further comparison for future work.

\vspace{-0.1cm}
\begin{proposition}
\label{prop:suff}
A sufficient condition for layer $\ell$ to be linear within the input-space region defined by $\mV$ (recall \cref{eq:V}) is
\vspace{-0.2cm}
\begin{align}
    0\leq \min_{(p,k) \in [P] \times [D^{(\ell+1)}]}\mH_{p,k}\vs_k\label{eq:b}
\end{align}
with $\mH \triangleq \mV^{(\ell)}(\mW^{(\ell)})^T + \mathbf{1}_{P}(\vb^{(\ell)})^T$ and with $\vs_k$ being $1$ if $\# \{i:(\mH)_{i,k}>0\}\geq P/2$ and $-1$ otherwise.
\end{proposition}
\vspace{-0.1cm}
In the above, $\mH$ is simply the pre-activation of layer $\ell$ given input $\mV^{(\ell)}$, and \cref{eq:b} ensures that all vertices have the same sign pattern as $\vs$; the latter determines on which side of the region $\mV^{(\ell)}$ each hyperplane is projected to.
Combining \cref{thm:police,prop:suff} leads to the POLICE algorithm given in \cref{algo:police}. Note that our choice of majority vote for $\vs$ is arbitrary, other options would be valid as well depending on the desired properties that POLICE should follow. For example, one could find that direction based on margins instead. We now turn to empirically validate it in a variety of tasks in the following section.

\begin{algorithm}[t!]
\caption{\small
POLICE procedure given a DNN $f_{\vtheta}$ (recall \cref{eq:DNN}) with activations $\sigma$ that can be (leaky-)ReLU or absolute value, and the region $R$ expressed by its vertices (recall \cref{eq:V}). POLICE strictly enforces that $f_{\vtheta}$ be affine in $R$; for clarity we denote by ${\rm MajorityVote}(., {\rm axis}=0)$ the $\vs$ vector from \cref{prop:suff}.}\label{algo:police}
\begin{center}
{\bf \underline{Train time procedure for each mini-batch $\mX$:}}
\end{center}\vspace{-0.3cm}
\begin{algorithmic}[1]
\Require $\mX \in \mathbb{R}^{N \times D},\mV\in \mathbb{R}^{V \times D}$
\State $\mV^{(1)} \triangleq \mV, \mX^{(1)} \triangleq \mX$\Comment{Initialization}
\For{$\ell=1,\dots,L$}\Comment{Forward-pass}
    \State $ \mH \leftarrow \mV^{(\ell)} (\mW^{(\ell)})^T + \mathbf{1}_{P}(\vb^{(\ell)})^T$
    \State $ \vs = {\rm MajorityVote}({\rm sign}(\mH), {\rm axis}=0)$
    \State $\vc = {\rm ReLU}(-\mH {\rm diag}(\vs)).{\rm max}({\rm axis}=0) \odot \vs$\label{lst:line:c}
    \State $\mX^{(\ell+1)}\leftarrow \sigma\left(\mX^{(\ell)}(\mW^{(\ell)})^T + \mathbf{1}_{N}(\vb^{(\ell)}+\vc)^T\right)$
    \State $\mV^{(\ell+1)}\leftarrow \sigma\left(\mV^{(\ell)}(\mW^{(\ell)})^T + \mathbf{1}_{P}(\vb^{(\ell)}+\vc)^T\right)$
\EndFor
\Ensure $\mX^{(L)}$\Comment{Evaluate loss and back-propagate as usual}
\end{algorithmic}\vspace{-0.3cm}
\begin{center}
{\bf \underline{Test time procedure (to run once post-training):}}
\end{center}\vspace{-0.3cm}
\begin{algorithmic}
\State Do the train time procedure but without $\mX,\mX^{(1)},\dots$ and set $\vb^{(\ell)} \leftarrow \vb^{(\ell)}+\vc,\forall \ell$ with $\vc$ from Line~\ref{lst:line:c}.
\end{algorithmic}
\end{algorithm}

\begin{figure*}[t!]
\begin{minipage}{0.8\linewidth}
    \hspace{1.15cm}
    \begin{minipage}{0.3\linewidth}
    \centering
    \underline{target function}
    \end{minipage}
    \begin{minipage}{0.3\linewidth}
    \centering
    \underline{POLICE constrained}\\
    \underline{DNN approximation}
    \end{minipage}\\
    \includegraphics[width=\linewidth]{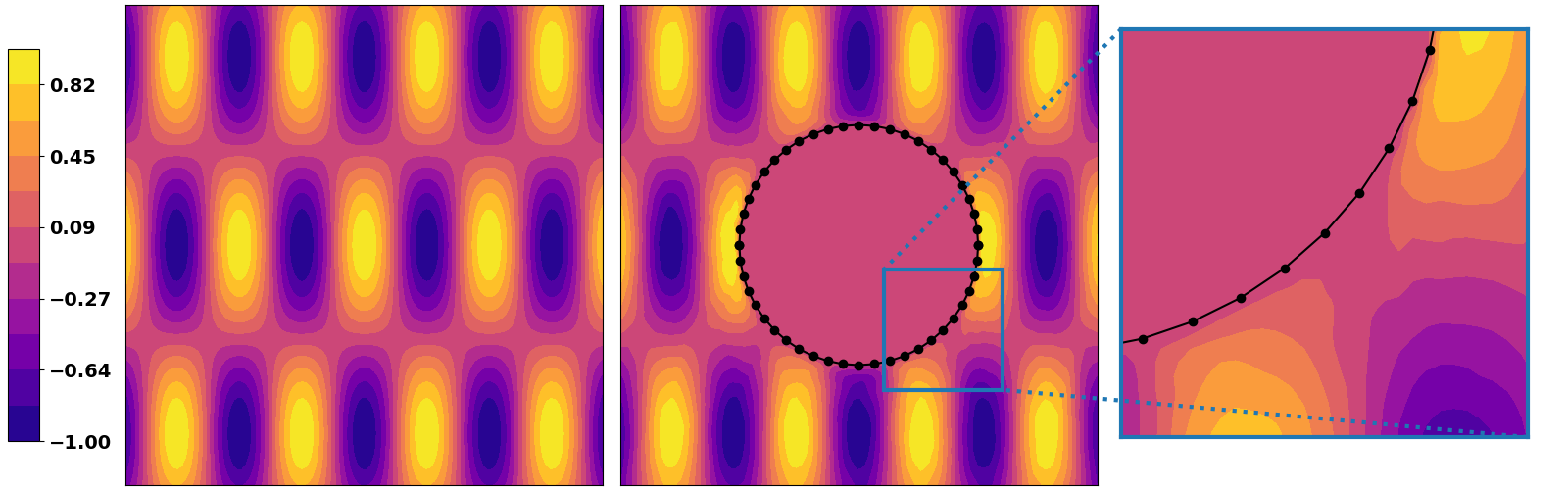}
    \includegraphics[width=\linewidth]{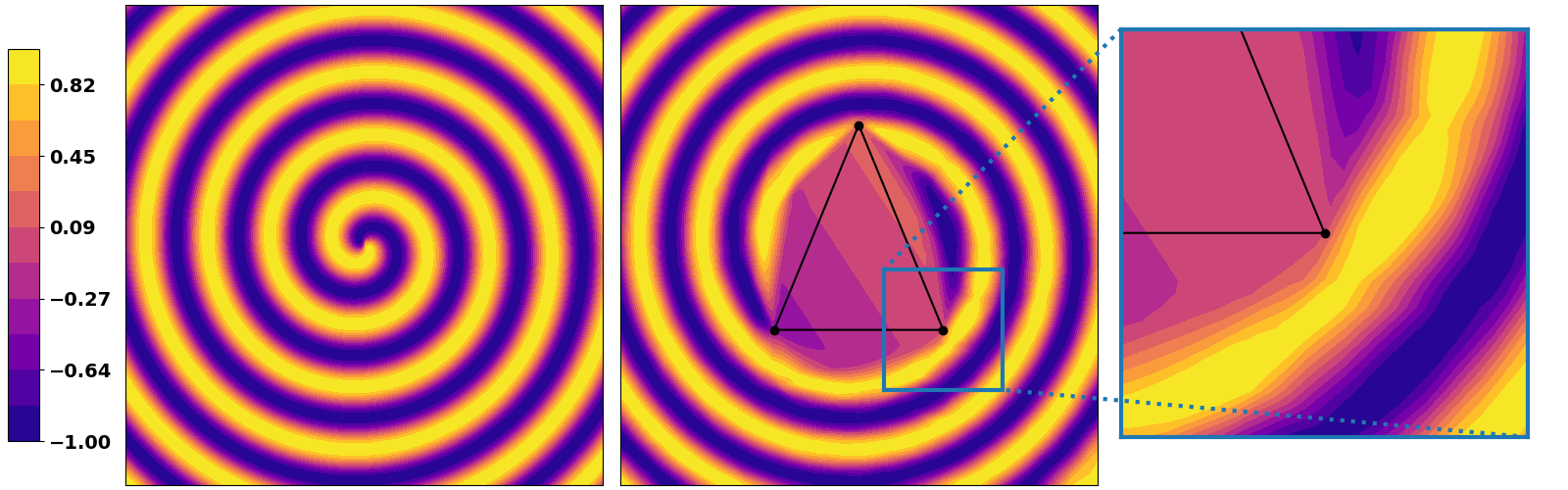}
    \end{minipage}
    \begin{minipage}{0.195\linewidth}
        \caption{\small Regression task of a $2$-dimensional space to a $1$-dimensional space with target function depicted in the {\bf left column} with leaky-ReLU MLP of depth $4$ and width $256$ and for which the POLICE constrains the DNN to be an affine mapping within the region $R$ delimited by the {\bf black lines} as depicted in the {\bf second column}. On the {\bf right column} a zoom-in is provided demonstrating how gradient based learning adapted the DNN parameters to very accurately approximate the target function.}
    \label{fig:cases}
    \end{minipage}
    \vspace{-0.7cm}
\end{figure*}

\begin{table}[t!]
\def\arraystretch{0.9}
\setlength\tabcolsep{2.6 pt}
    \small
    \caption{\small Average (over $1024$ runs) time (in ms.) and std to compute a forward-pass and a backward-pass with gradient accumulation in a standard DNN without any constraint enforced versus a POLICEd DNN for various input dimensions, widths and depths; given a mini-batch size of $1024$. We recall that at test time a POLICEd DNN has zero overhead (recall \cref{algo:police}).}
    \label{tab:times}
    \centering
    \begin{tabular}{c|c|c|c|c|c|c}
        input dim. $D$                 & 2              & 2             & 2              & 784           & 784            & 3072 \\
        depth $L$                      & 2.             & 4             & 4              & 2             & 8              & 6 \\
        widths $D^{(\ell)}$& 256            & 64            & 4096           & 1024          & 1024           & 4096 \\ \toprule
       no constr.                          & 0.9$\pm$0   & 1.3$\pm$0 & 27.9$\pm$3 & 0.9$\pm$0 & 3.1$\pm$0   & 51.9$\pm$5 \\ 
       POLICEd                       & 4.3$\pm$0   & 7.7$\pm$1 & 40.3$\pm$1 & 5.2$\pm$0 & 20.4$\pm$1 & 202.2$\pm$12 \\
       slow-down                         & $\times$4.5    & $\times$ 5.7  & $\times$ 1.4   & $\times$ 5.5  & $\times$ 6.6   & $\times$ 3.9 \\ \bottomrule
    \end{tabular}
    \vspace{-0.2cm}
\end{table}
 
\vspace{-0.2cm}
\subsection{Empirical Validation}
\label{sec:validation}
\vspace{-0.2cm}

\begin{figure}[t!]
    \centering
    \begin{minipage}{0.17\linewidth}
    \centering
    \underline{target function}
    \end{minipage}
    \begin{minipage}{0.82\linewidth}
    \centering
    \underline{POLICE constrained DNN}\\
    \begin{minipage}{0.32\linewidth}
        \hspace{1.2cm}
        step $5$
    \end{minipage}
    \begin{minipage}{0.32\linewidth}
        \hspace{0.9cm}
        step $50$
    \end{minipage}
    \begin{minipage}{0.32\linewidth}
        \centering
        step $10000$
    \end{minipage}
    \end{minipage}
    \includegraphics[width=\linewidth]{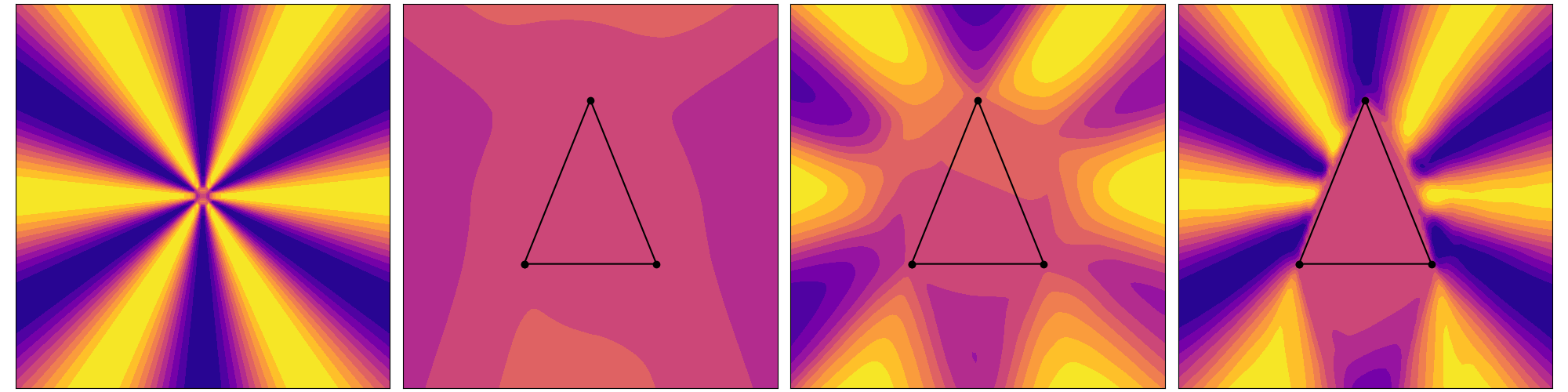}
    \vspace{-0.8cm}
    \caption{\small Evolution of the DNN function approximation to the target function ({\bf left}) at different training steps (5,50,10000) in each {\bf column}. At any point during training POLICE strictly enforced the constraints on the region $R$ delimited by the {\bf black lines} in a differentiable way to allow gradient based learning to discover a good value for the DNN parameters.}
    \label{fig:evolution}
    \vspace{-0.4cm}
\end{figure}

We now propose to empirically validate the proposed POLICE method from \cref{algo:police}.
First, we point out that POLICE can be applied regardless of the downstream task that one aims at solving e.g. classification or regression as presented in \cref{fig:classification,fig:cases} respectively. In either case, the exact same methodology is employed (\cref{algo:police}). To complement the proposed classification case, we also provide in \cref{fig:evolution} the evolution of the DNN approximation at different training steps. We observe how the POLICE method enable gradient based learning to take into account the constraints so that the approximation can be of high-quality despite the constraint enforcement. Crucially, {\em POLICE provides a strict constraint enforcement at any point during training requiring no tuning or sampling}. We provide training times in \cref{tab:times} for the case with $R$ being the simplex of the considered ambient space, and recall the reader that POLICE has zero overhead at test time. We observe that especially as the dimension or the width of the model grows, as POLICE is able to provide relatively low computational overhead. In any case, the computational slow-down seems to lie between $1.4$ and $6.6$ in all studied cases. Additionally, those numbers were obtained from a naive Pytorch reproduction of \cref{algo:police} and we expect the slow-down factor to greatly diminish by optimizing the implementation which we leave for future work.

\vspace{-0.3cm}
\section{Conclusion and Future Work}
\label{sec:conclusion}
\vspace{-0.2cm}

We proposed a novel algorithm --POLICE-- to provably enforce affine constraints into DNNs that employ continuous piecewise affine nonlinearities such as (leaky-)ReLU, absolute value or max-pooling. Given polytopal region $R$, POLICE enforces the DNN to be affine on $R$ only by a forward-pass through the DNN of the vertices defining $R$ (\cref{algo:police}) making it computationally efficient (\cref{tab:times}) and lightweight to implement (no change in the model or optimizer). We believe that strict enforcement of the affine constraint is key to enable critical applications to leverage DNNs e.g. to enforce \cref{eq:constraint,eq:constraint2}.
Among future directions we hope to explore the constraint enforcement on multiple regions simultaneously, and the extension to DNN employing smooth activation functions using a probabilistic argument as done e.g. in \cite{balestriero2018hard,humayun2022polarity}. Regarding the application of POLICE, we believe that implicit representation DNNs \cite{sitzmann2020implicit} could be fruitful beneficiaries.

\clearpage
\newpage

\bibliographystyle{IEEE}
\bibliography{bibliography}

\end{document}